\documentclass[runningheads]{llncs}
\usepackage{graphicx}

\usepackage{tikz}
\usepackage{comment} 
\usepackage{amsmath,amssymb} 
\usepackage{color}
\usepackage{bm}
\usepackage{multirow}
\usepackage[caption=false]{subfig}
\usepackage{microtype}
\usepackage[american]{babel}
\newcommand{\ie}{\emph{i.e.}}

\begin{document}
\pagestyle{headings}
\mainmatter
\def\ECCVSubNumber{100}  

\title{Hierarchical Context Embedding for Region-based Object Detection}

\titlerunning{Hierarchical Context Embedding for Region-based Object Detection}
%
\author{Zhao-Min Chen\inst{1, 2} \and
Xin Jin\inst{2} \and
Borui Zhao\inst{2} \and
Xiu-Shen Wei\inst{2,*} \and
Yanwen Guo\inst{1,*}
}
\renewcommand{\thefootnote}{*}
\footnotetext[1]{Corresponding authors}
\authorrunning{Z.-M. Chen, X. Jin, B. Zhao, X.-S. Wei, and Y. Guo}
%
\institute{State Key Laboratory for Novel Software Technology, Nanjing University, China \and
Megvii Research Nanjing, Megvii Technology, China\\
{\tt\small \{chenzhaomin123, weixs.gm\}@gmail.com, \{jinxin, zhaoborui\}@megvii.com, ywguo@nju.edu.cn}\\
}
\maketitle

\begin{abstract}
State-of-the-art two-stage object detectors apply a classifier to a sparse set of object proposals, relying on region-wise features extracted by RoIPool or RoIAlign as inputs. The region-wise features, in spite of aligning well with the proposal locations, may still lack the crucial context information which is necessary for filtering out noisy background detections, as well as recognizing objects possessing no distinctive appearances. To address this issue, we present a simple but effective Hierarchical Context Embedding (HCE) framework, which can be applied as a plug-and-play component, to facilitate the classification ability of a series of region-based detectors by mining contextual cues. Specifically, to advance the recognition of context-dependent object categories, we propose an image-level categorical embedding module which leverages the holistic image-level context to learn object-level concepts. Then, novel RoI features are generated by exploiting hierarchically embedded context information beneath both whole images and interested regions, which are also complementary to conventional RoI features. Moreover, to make full use of our hierarchical contextual RoI features, we propose the early-and-late fusion strategies (\ie, feature fusion and confidence fusion), which can be combined to boost the classification accuracy of region-based detectors. Comprehensive experiments demonstrate that our HCE framework is flexible and generalizable, leading to significant and consistent improvements upon various region-based detectors, including FPN, Cascade R-CNN and Mask R-CNN.
\end{abstract}

\section{Introduction}

The region-based object detectors~\cite{rcnn,fastrcnn,fasterrcnn,fpn,cascade,ecr} popularized by R-CNN framework~\cite{rcnn} are conceptually intuitive and flexible, and have achieved top accuracies on challenging benchmarks like MS-COCO~\cite{coco}. Region-based detectors first generate a sparse set of object proposals, and then refine the proposal locations and classify them as one of the foreground classes or as background using a detection head. One crucial module in such a proposal-driven pipeline is the RoIPool~\cite{fastrcnn} or RoIAlign~\cite{maskrcnn} operator, which is responsible for extracting RoI (Region of Interests) features aligned with the proposal locations for the detection head.

\begin{figure}[t]
	\centering
	\subfloat[Filtering out noisy detections.] {\includegraphics[width=0.48\columnwidth]{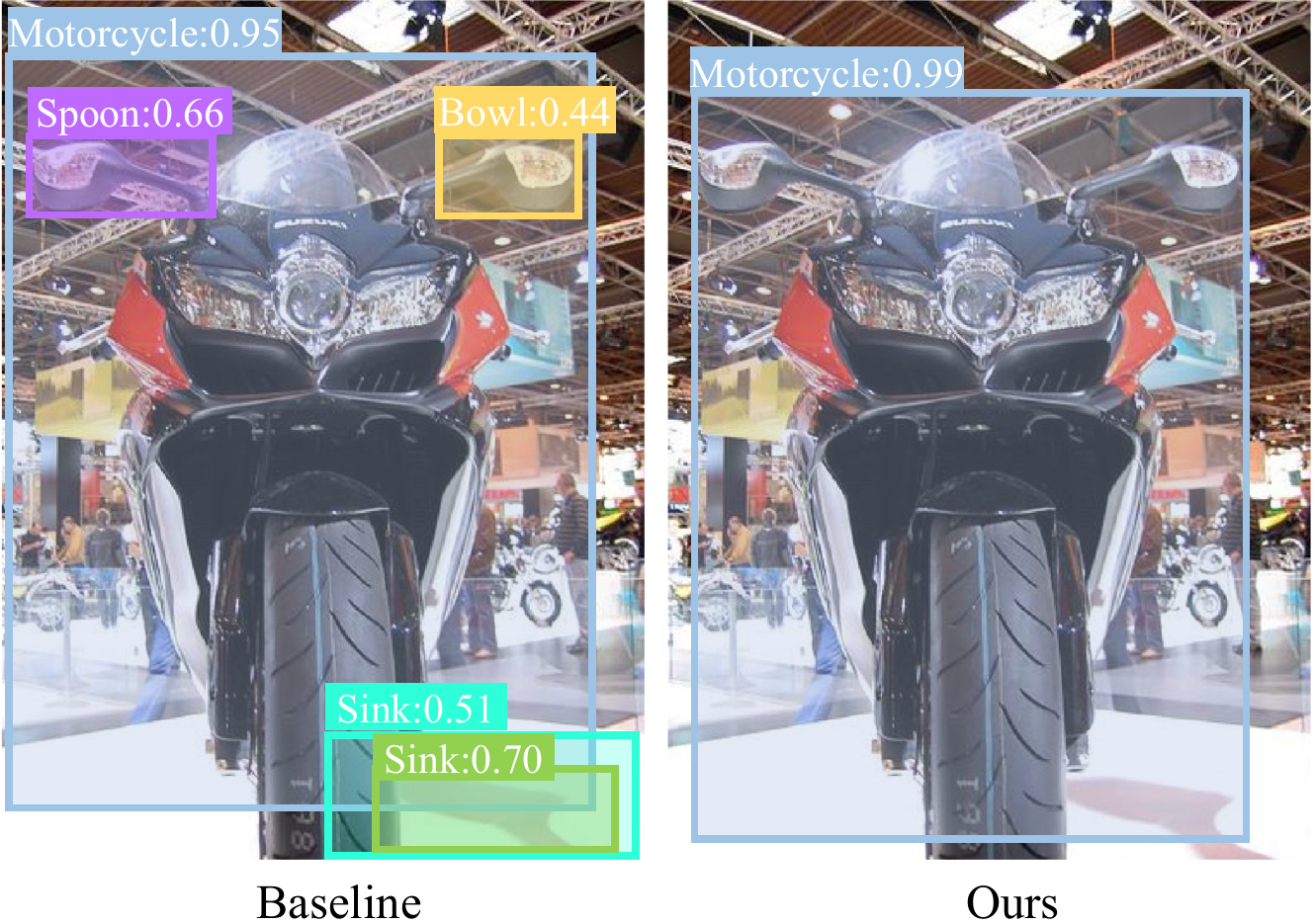} }
	\subfloat[Recognizing indistinctive objects.]  {\includegraphics[width=0.48\columnwidth]{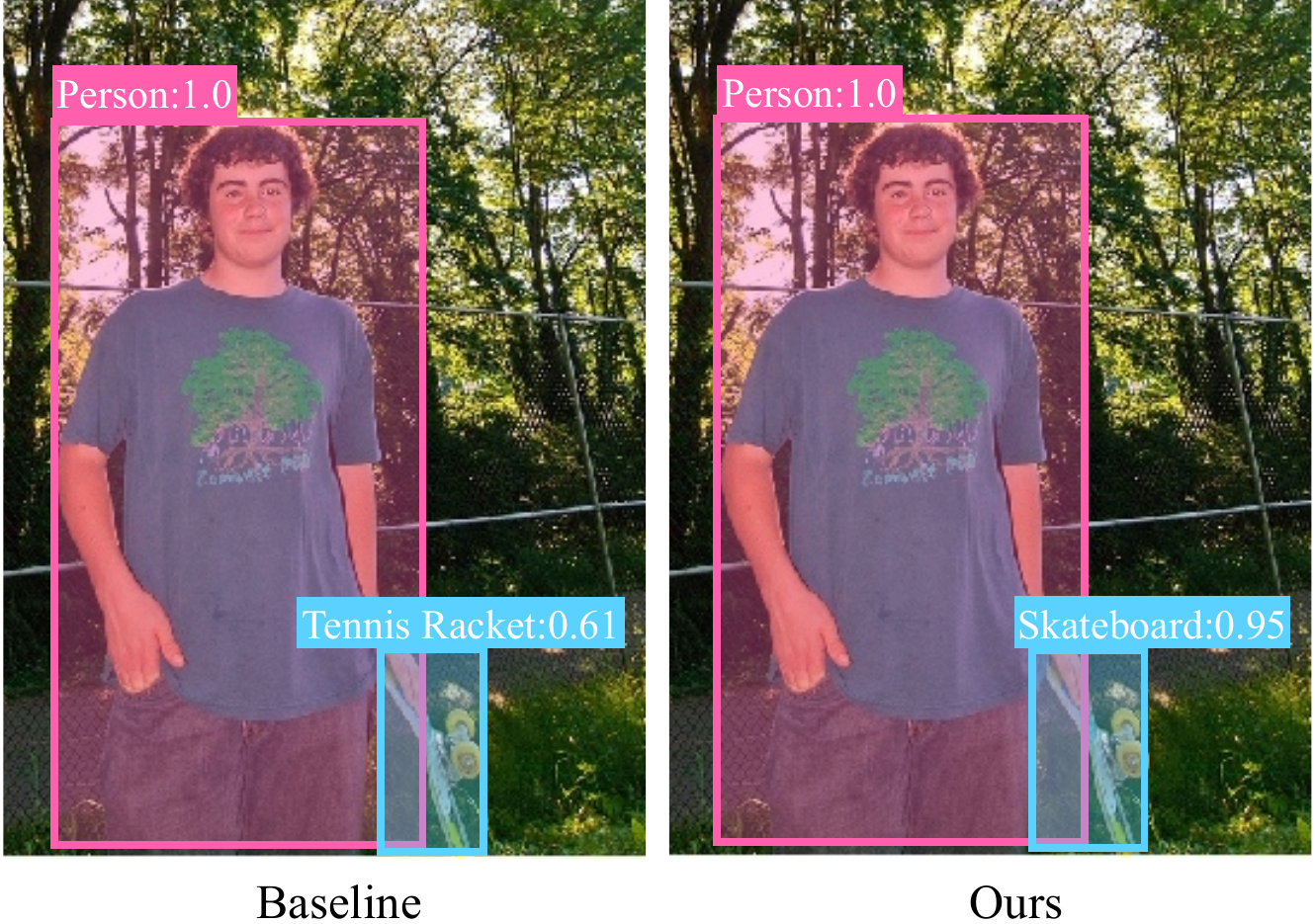} }
	\scriptsize
	\caption{Motivation and example results of our Hierarchical Context Embedding (HCE) framework. By incorporating discriminative context information, our framework can effectively filter out the noisy false positive background detections, and correctly classify objects (\textit{e.g.}, ``\texttt{skateboard}'') which possess no distinctive appearances.}
	\label{fig:motivation}
\end{figure}

In this paper, we revisit the RoI features in region-based detectors from the perspective of context information embedding. Our key motivation relies on the fact that while each RoI in very deep CNNs may have a very large theoretical receptive field that often spans the entire input image~\cite{fastrcnn}. However, the effective receptive field~\cite{Luo2016} may only occupy a fraction of the full theoretical receptive field, making the RoI features insufficient for characterizing objects that are highly dependent on context information, such as ``\texttt{bowl}'', ``\texttt{skateboard}'' etc. Here, the contextual information means any auxiliary information that can assist in suppressing the false positive detections in noisy backgrounds, or recognizing objects that have no distinctive appearances themselves. For example, as shown in~Fig.~\ref{fig:motivation} (a), the semantic features of ``\texttt{motorcycle}'' are strong evidences for filtering out the activations of irrelevant object categories like ``\texttt{spoon}'', ``\texttt{bowl}'', and ``\texttt{sink}''. On the other hand, as shown by~Fig.~\ref{fig:motivation} (b), the scene and even the human pose are useful clues for correctly classifying a proposal as ``\texttt{skateboard}'', rather than ``\texttt{tennis racket}''.

Recently, several works exploited the region-level context information to improve the localization ability of two-stage detectors. Chen~\textit{et al.}~\cite{context} demonstrated that rich contextual information from neighboring regions can better refine the proposal locations for two-stage detectors. Kantorov~\textit{et al.}~\cite{kantorov2016contextlocnet} leveraged the surrounding context regions to improve weakly supervised object localization. However, to the best of our knowledge, currently there is no enabling framework which is systematically designed for embedding context information to improve the \textit{classification ability} of region-based detectors.

In this paper, we present a novel Hierarchical Context Embedding (HCE) framework for region-based object detectors. Our framework consists of three modules. Firstly, we consider that the simplest way to break the contextual limit in object detection, is to partially cast the object-level feature learning as an image-level multi-label classification task. Building upon this realization, we design an image-level categorical embedding module, which in essence is a multi-label classifier upon the detection backbone, in parallel with the existing region-based detection head. It enables the backbone to exploit the whole image context to learn discriminative features for context-dependent object categories. Even as a standalone enhancement, our image-level categorical embedding module can lead to improvements over existing region-based detectors.

Upon the image-level categorical embedding module, at the instance-level, we design a simple but effective process to generate hierarchical contextual RoI features which can be directly utilized by the region-wise detection head. Because our contextual RoI features are enhanced by image-level categorical supervisions and exploit larger contexts, they are by nature complementary to conventional RoI features, which is trained by region-based detectors and only exploits limited context. Later, the early-and-late strategies, \ie, feature fusion and confidence fusion, are designed to make full use of our contextual RoI features. By quantitative experiments, we demonstrate that they can be combined to further boost the classification accuracy of the detection head.

In general, our proposed HCE framework is easy to implement and is end-to-end trainable. We conduct extensive experiments on MS-COCO 2017~\cite{coco} to validate the effectiveness of our HCE framework. Without bells and whistles, our HCE framework delivers consistent accuracy improvements for almost all existing mainstream region-based detectors, including FPN~\cite{fpn}, Mask R-CNN~\cite{maskrcnn} and Cascade R-CNN~\cite{cascade}. We also conduct ablation studies to verify the effectiveness of each module involved in our HCE framework. Fig.~\ref{fig:motivation} gives the example images of the baseline method and our method, which demonstrates that our framework can effectively filter out the noisy background detections and correctly classify indistinctive objects by leveraging the context information it exploited.

\section{Related Work}

\subsection{Region-based Object Detection} 
Convolutional neural networks have lead to a paradigm shift of object detection in the past decades~\cite{liu2019deep}. Among a large number of approaches, the two-stage R-CNN series~\cite{rcnn,fastrcnn,fasterrcnn,fpn,cascade} have become the leading detection framework. The pioneer work, \ie, R-CNN~\cite{rcnn}, extracts region proposals from image with selective search~\cite{uijlings2013selective}, and applies a convolutional network to classify each region of interests independently. Fast R-CNN~\cite{fastrcnn} improves R-CNN by sharing convolutional features among RoIs, which enables fast training and inference. Then, Faster R-CNN~\cite{fasterrcnn} advances the region proposal generation with a Region Proposal Network (RPN). RPN shares the feature extraction backbone with the detection head, which in essence is a Fast R-CNN~\cite{fasterrcnn}. Faster R-CNN is a famous two-stage detection framework, and is the foundation for many follow-up works~\cite{dai2016r,fpn}.

Over very recent years, several algorithms have been proposed to further improve the two-stage Faster R-CNN framework. For example, Feature Pyramid Networks (FPN)~\cite{fpn} constructed inherent CNN feature pyramids, which can largely improve the detection performance of small objects. Mask R-CNN~\cite{maskrcnn} extended Faster RCNN by constructing the mask branch, and boosted the performance of both object detection and instance segmentation. Cascade R-CNN~\cite{cascade} utilized multi-stage training strategy to progressively improve the quality of region proposals, and demonstrated significant gains for high quality (measured by higher IoUs) object detection. Complementary to these works, in this paper, we focus on developing a Hierarchical Context Embedding (HCE) framework to improve the \textit{classification ability} of all region-based detectors. Thanks to the simplicity and generalization ability of our HCE framework, it brings consistent and significant improvements over aforementioned leading region-based detectors, \textit{e.g.}, FPN, Mask R-CNN and Cascade R-CNN.

\subsection{Context Information for Object Detection}
In object detection, both global context~\cite{galleguillos2010context} and local context~\cite{rabinovich2007objects} are widely exploited for improving performance, especially when object appearances are insufficient due to small object size, occlusion, or poor image quality. Our work is inspired by some of previous works, but the key motivation or implementation significantly differ with these works. Next, we review several topics in object detection, which are closely related to our work.

\subsubsection{Combined Localization and Classification.} Before the era of deep learning, Harzallah \textit{et al.}~\cite{harzallah2009combining} proposed to combine two closely related tasks, \textit{i.e.}, object localization and image classification. They demonstrated that classification can improve detection by a contextual combination and vice versa. Similar in spirit, we utilize the fully image-level context to learn object-level concepts. But differently, we utilize global context to learn CNN features rather than hand-crafted features adopted in~\cite{harzallah2009combining}. Furthermore, we integrate hierarchical contextual clues beneath both whole images and interested regions to modern region-based CNN detectors, rather than the traditional sliding window detector used by~\cite{harzallah2009combining}.

\subsubsection{Region Proposal Refinement.} Recently, Chen \textit{et al.}~\cite{context} explored the rich contextual information to refine the region proposals for object detection. The neighboring regions with useful contexts can benefit the localization quality of region proposals, which further lead to better detection performance. Instead of refining proposals, we focus on improving the \textit{classification ability} of region-based detectors by embedding hierarchical contextual clues.

\subsubsection{Weakly-Supervised Object Detection.} 
In weakly supervised object detection, the bounding box annotations are not provided, and only image-level categorical labels are available. The common practice~\cite{cinbis2016weakly,weakly,bilen2015weakly,kantorov2016contextlocnet} in this area is to first generate a set of noisy object proposals, and then learn from these noisy proposals with specially designed robust algorithms. Among them, Kantorov \textit{et al.}~\cite{kantorov2016contextlocnet} proposed a context-aware deep network which leverages the surrounding context regions to improve localization. Unlike the usage of region-level context information~\cite{kantorov2016contextlocnet} for weakly supervised detection, we focus on the task of fully-supervised object detection, and particularly exploit global image-level context to advance the recognition of context-dependent object categories.

\subsection{Context Information for Other Vision Tasks}
Beyond object detection, context information has also been utilized to improve other vision tasks. For example, Wang \textit{et al.}~\cite{rnn_attention} leveraged attention mechanisms and LSTMs to discover semantic-aware regions and capture the long-range contextual dependencies for multi-label image recognition. He \textit{et al.}~\cite{adaptive} proposed an adaptive context module to generate multi-scale context representations for semantic segmentation. Qu \textit{et al.}~\cite{deshadownet} embedded multi-context information (the appearance of the input image and semantic understanding) to obtain the shadow matte. Byeon \textit{et al.} ~\cite{contextvp} leveraged the LSTM units to capture the entire available past context on video prediction. Li \textit{et al.}~\cite{derain}  adopted the dilated convolution to acquire more contextual information for single image deraining.

\section{Approach}

\subsection{Framework Overview}
\begin{figure}[t]
	\centering
	\includegraphics[width=1.0\textwidth]{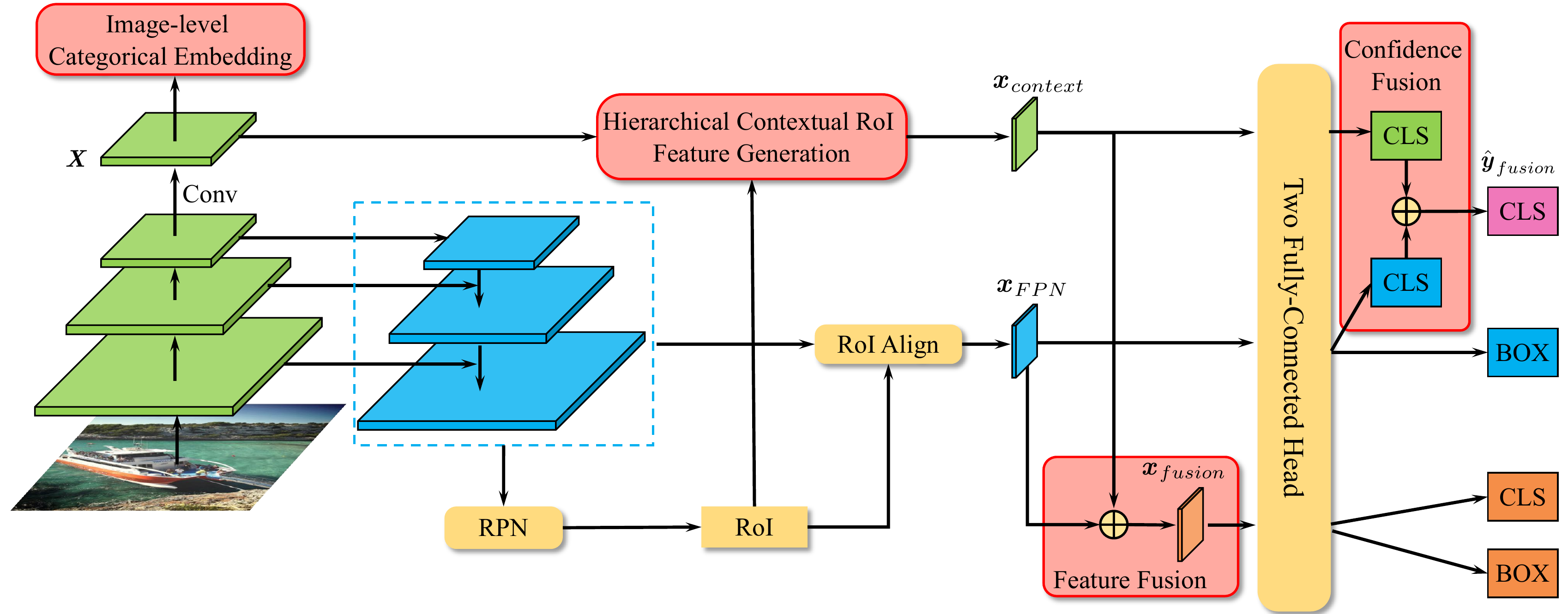}
	\caption{Overview of our Hierarchical Context Embedding (HCE) framework. 
	At the image-level, we design an \emph{image-level categorical embedding} module upon the detection backbone, which enables the network to learn object-level concepts from global-level context. At the instance-level, we generate \emph{hierarchical contextual RoI features} that are complementary to conventional RoI features, and design the early-and-late fusion strategies (\ie, \emph{feature fusion} and \emph{confidence fusion}) to make full use of the contextual RoI features for improving the classification accuracy of the detection head.
	}
	\label{fig:model}
\end{figure}

We begin by briefly describing our Hierarchical Context Embedding (HCE) framework (see Fig.~\ref{fig:model}) for region-based object detection. Firstly, an image-level categorical embedding module is employed to advance the feature learning of the objects that are highly dependent on larger context clues. Then, hierarchical contextual RoI features are generated by fusing both instance-level and global-level information derived from the image-level categorical embedding module. Finally, early-and-late fusion modules are designed to make full use of the contextual RoI features to improve the classification performance. Our HCE framework is flexible and generalizable, as it can be applied as a plug-and-play component for almost all mainstream region-based object detectors. 

\subsection{Image-Level Categorical Embedding}

As aforementioned, conventional RoI-based training for region-based detectors may lack the context information, which is crucial for learning discriminative filters for context-dependent objects. To break this limitation, in parallel with the RoI-based branch, we exploit image-level categorical embedding upon the detection backbone, enabling the backbone to learn object-level concepts adaptively from \textit{global-level} context. 
Our image-level categorical embedding module does not require additional annotations, as the image-level labels can be conveniently obtained by collecting all instance-level categories in an image.

\begin{figure}[t]
	\centering
	\subfloat[Image-level categorical embedding.] {\includegraphics[width=0.48\columnwidth]{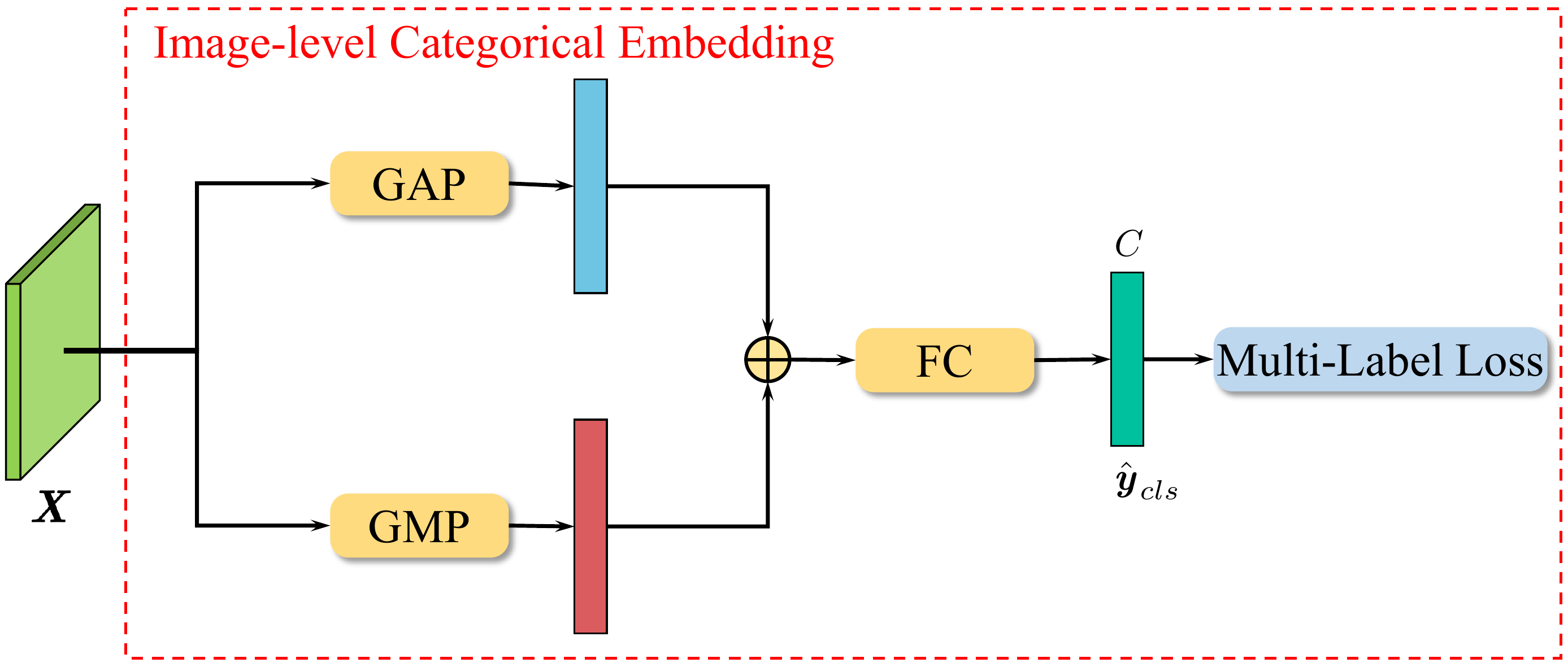} }
	\subfloat[Hierarchical contextual RoI feature.]  {\includegraphics[width=0.48\columnwidth]{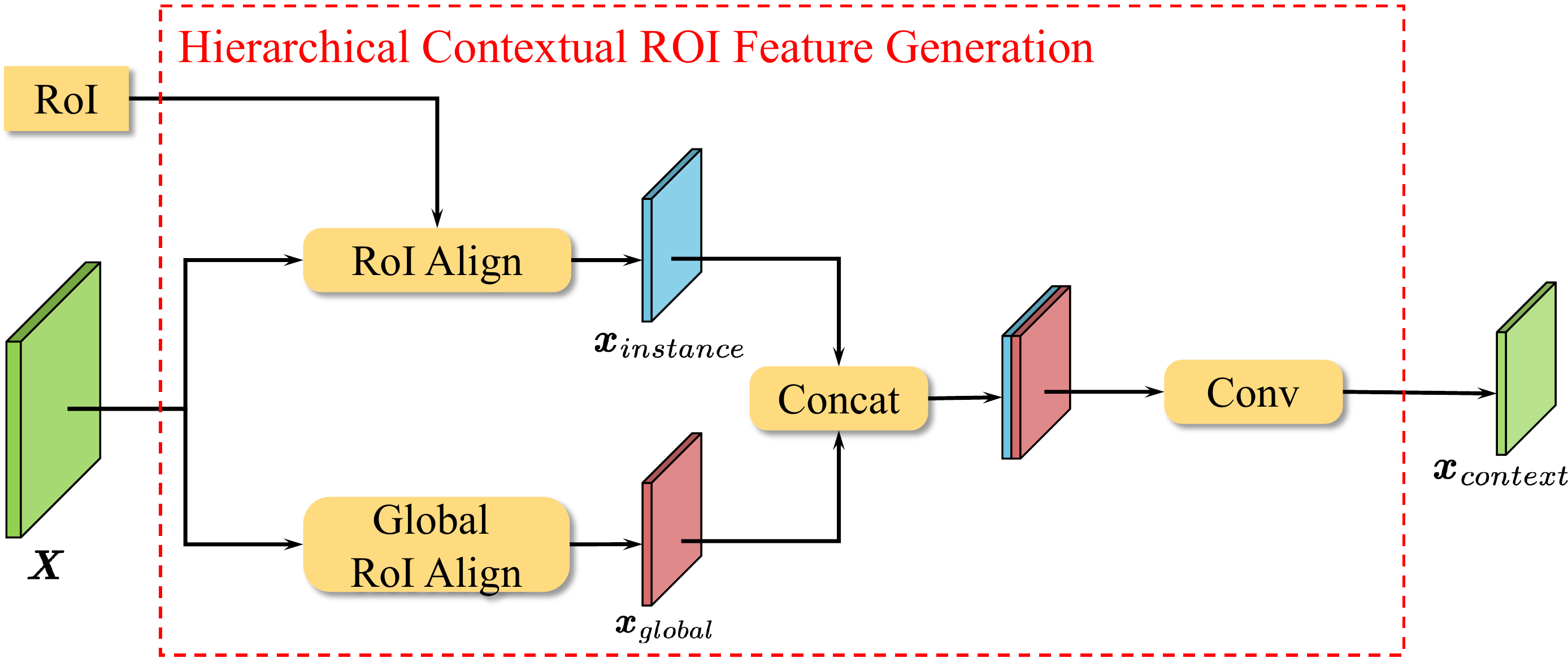} }
	\caption{The design of our image-level categorical embedding module and hierarchical contextual RoI feature generation module.}
	\label{fig:module}
\end{figure}

Essentially, our image-level categorical embedding module is based on a multi-label classifier. As shown in Fig.~\ref{fig:model} and Fig.~\ref{fig:module} (a), we first apply a $3 \times 3$ convolution layer on the output of ResNet conv$_5$ to obtain the input feature map, and then employ both global max-pooling (GMP) and global average-pooling (GAP) for feature aggregation (as in~\cite{woo2018cbam}). Here, the additional $3 \times 3$ convolution layer aims to alleviate the possible slide effects over the original detection backbone.  

We refer to the input feature map to our image-level embedding module as \emph{context-embedded image feature}. This is because the input feature map conveys whole image context for learning all object categories that appear in the image, and in turn, each location of the feature map is supervised by all object categories. By contrast, conventional RoI-based trained by region-based detectors only exploits limited context for learning each object category.

Formally, let $\bm{X} \in \mathbb{R}^{d \times h \times w}$ denote the input feature map, where $d$ is the channel dimensionality, $h$ and $w$ are the height and width, respectively. Then, the multi-label classifier is constructed by $C$ binary classifiers for all categories:
\begin{equation}
	\hat{\bm{y}}_{cls} = f_{cls}((f_{gmp}(\bm{X}) + f_{gap}(\bm{X}))) \in \mathbb{R}^{C} \,,
\end{equation}
where $C$ denotes the number of categories, each element of $\hat{\bm{y}}_{cls}$ is a confidence score (logits), and $f_{cls}$ is binary classifier modeled as one fully-connected layer. We assume that the ground truth label of an image is $\bm{y} \in \mathbb{R}^{C}$, where $y^{i} = \{0, 1\}$ denotes whether object of category $i$ appears in the image or not. The multi-label loss can be formulated as follows

\begin{equation}
	\mathcal{L}_{mll}= - \sum_{c=1}^{C}y^{c}\log(\sigma(\hat{y}^{c}_{cls})) + (1-y^{c})\log(1-\sigma(\hat{y}^{c}_{cls})) \,,
\end{equation}
where $\sigma(\cdot)$ is the sigmoid function.

Because the global feature learning strategy is complementary to RoI-based training, our image-level categorical embedding module standalone can boost the performance of existing region-based detectors (demonstrated later by experiments, cf. Table~\ref{table:context}). However, one limitation of image-level categorical embedding might be that the derived context-embedded image feature can not be directly used by the detection head.

\subsection{Hierarchical Contextual RoI Feature Generation}

To further benefit region-wise classification, we generate hierarchical contextual RoI features by combining the instance-level and global-level information from the context-embedded image features. The hierarchical contextual RoI feature generation process is shown in Fig.~\ref{fig:module} (b).

\subsubsection{Context-Embedded Instance-Level Feature}

We apply RoIAlign~\cite{maskrcnn} with proposals generated by RPN on the context-embedded feature map $\bm{X}$ to obtain RoI features $\bm{x}_{instance}$:
\begin{equation}
	\bm{x}_{instance} = f_{RoIAlign}(\bm{X}; h', w') \in \mathbb{R}^{d \times 7 \times 7} \,,
	\label{eq:instance}
\end{equation}
where $f_{RoIAlign}(\cdot)$ is the RoIAlign operation and $h'$ and $w'$ are the height and width of the RoI, respectively. As $\bm{x}_{instance}$ is extracted from the context-embedded image feature $\bm{X}$, we term it as \emph{context-embedded instance-level feature}.

\subsubsection{Context-Aggregated Global-Level Feature} To leverage larger context, we exploit RoIAlign on the context-embedded image feature $\bm{X}$ to aggregate the global-level context. We refer to the derived RoI feature as \emph{context-aggregated global-level feature} $\bm{x}_{global}$:
\begin{equation}
	\bm{x}_{global} = f_{RoIAlign}(\bm{X}; H, W) \in \mathbb{R}^{d \times 7 \times 7} \,,
	\label{eq:global}
\end{equation}
where $H$ and $W$ are the height and width of the input image, respectively.

Once context-embedded instance-level feature $\bm{x}_{instance}$ and context-aggregated global-level feature $\bm{x}_{global}$ obtained, we concatenate these two RoI features and apply a $1 \times 1$ convolution layer to obtain our hierarchical contextual RoI feature $\bm{x}_{context}$:
\begin{equation}
	\bm{x}_{context} = f_{conv}([\bm{x}_{instance} : \bm{x}_{global}]) \in \mathbb{R}^{d \times 7 \times 7} \,,
\end{equation}
where $f_{conv}(\cdot)$ denotes the $1 \times 1$ convolution operation, $[:]$ refers to concatenation and the ReLU nonlinearity operations are performed following the convolution layer. As the resulting hierarchical contextual RoI feature $\bm{x}_{context}$ absorbs rich context information from the context-embedded image feature $\bm{X}$, it is by nature complementary to the conventional RoI feature extracted from the feature pyramid network (FPN)~\cite{fpn}.

\subsection{Early-and-Late Fusion and Inference}

To make full use of our contextual RoI feature $\bm{x}_{context}$, we design the early-and-late fusion strategies, \ie, feature fusion and confidence fusion, which has been proven effective in many applications~\cite{gunes2005affect,ebersbach2017fusion}. We show that early-and-late fusion is also well suited to improve region-wise detectors, as it can fully absorb hierarchically embedded information from different levels.

\subsubsection{Feature Fusion} To incorporate our contextual RoI features $\bm{x}_{context}$ into region-based detection pipeline, the simplest way is fusing them with the original RoI features extracted from the feature pyramid network (FPN)~\cite{fpn} with element addition. Formally, let $\bm{x}_{FPN}$ denote the original RoI feature extracted from FPN, and $\bm{x}_{fusion}$ denote the fused RoI feature, then we have:
\begin{equation}
	\bm{x}_{fusion} = \bm{x}_{context} + \bm{x}_{FPN} \in \mathbb{R}^{d \times 7 \times 7}.
\end{equation}

As shown in Fig.~\ref{fig:model}, the fused feature map $\bm{x}_{fusion}$ is then fed into the 2$fc$ detection head to produce refined bounding boxes and classification scores.

\subsubsection{Confidence Fusion} We also consider a simple confidence fusion strategy which is complementary to feature fusion. We apply the 2$fc$ head on our hierarchical contextual RoI feature $\bm{x}_{context}$ to produce a classification confidence (logits), and then fuse it with that from the corresponding FPN RoI feature $\bm{x}_{FPN}$ by addition. Formally, let $\hat{\bm{y}}_{fusion}$ denote the fused the confidence:
\begin{equation}
	\hat{\bm{y}}_{fusion} = f_{2fc}(\bm{x}_{context}) + f_{2fc}(\bm{x}_{FPN}) \in \mathbb{R}^{C}.
\end{equation}
The fused confidence is transformed by a soft-max layer to produce a novel classification score. 

For each proposal, the classification score $\hat{\bm{y}}_{fusion}$, paired with the refined bounding box predicted the FPN RoI feature, forms another prediction in parallel with the prediction from the feature fusion branch. It is worth mentioning that the weights of the 2$fc$ head applied on different RoI features are shared.

\subsubsection{Inferences} Our early-and-late fusion strategy produces two different predictions for a single object proposal. To obtain the final result, as shown by the pipeline in Fig.~\ref{fig:model}, we firstly collect all the boxes and confidences from two prediction branches (\ie, feature fusion and confidence fusion), and then perform NMS over all these boxes. Furthermore, as demonstrated later in experiments, while our two fusion strategies are complementary during training, using only one prediction branch during inference will not cause obvious performance drop but reduce computational cost. However, the performance by only using one fusion strategy for training is inferior to that by using two fusion strategies together.

\subsubsection{Loss Function}
The whole network is trained end-to-end, and the overall loss is computed as follows:
\begin{equation}
	\mathcal{L} = \mathcal{L}_{feat} + \mathcal{L}_{conf} + \mathcal{L}_{mll} + \mathcal{L}_{rpn} \,,
\end{equation}
where $\mathcal{L}_{feat}$ and $\mathcal{L}_{conf}$ are the losses for the feature fusion and confidence fusion branches, respectively. All loss terms are considered equally important, without extra hyper-parameters to characterize the trade-off between them, which reveals HCE is generalized and not tricky.

\section{Experiments}
We conduct extensive experiments on the MS-COCO 2017 dataset~\cite{coco} to demonstrate the effectiveness and generalization ability of our hierarchical context embedding framework. 
MS-COCO 2017 is the most popular benchmark for general object detection, which contains 80 object categories, 118K images for training, 5K images for validation (\texttt{val}) and 20K for testing (\texttt{test-dev}). 
We report the standard COCO-style Average Precision (AP) with different IoU thresholds from 0.5 to 0.95 with an interval of 0.05 as metric. All models are trained on COCO training set and evaluate on the \texttt{val} set. For fair comparisons with the state-of-the-art, we also report the results on the \texttt{test-dev} set.

\subsection{Implementation Details}

We implement our method and re-implement all baseline methods based on MMDetection codebase~\cite{mmdetection}. The re-implementations of the baselines strictly follow the default settings of MMDetection. Images are resized such that the short edge has 800 pixels while the long edge has less than 1333 pixels. We use no data augmentation except horizontal flipping for training. The ResNet is exploited as backbone, which is pre-trained on ImageNet~\cite{imagenet}. Models are trained in a batch size of 16 on 8 GPUs. We train all models with SGD optimizer for 12 epochs in the total, with the initial learning rate as 0.02 and decreased by a factor of 0.1 at 8th epoch and 11th epoch. Weight decay and momentum are set as 0.0001 and 0.9, respectively. We also adopt the linear warming up strategy to begin the training of our model.

\subsection{Comparisons with Baselines}

\begin{table}[t]
\centering
\setlength{\tabcolsep}{1.3mm}
\caption{Compared with baselines (FPN~\cite{fpn}, Mask R-CNN~\cite{maskrcnn} and Cascade R-CNN~\cite{cascade}) on MS-COCO 2017 \texttt{val}. ``HCE'' denotes that the models are trained and inferred on both feature fusion and confidence fusion. 
Clearly, our HCE framework achieves consistent accuracy gains overall all baseline detectors on all evaluation metrics.
}
\begin{tabular}{c|c|c|ccc|ccc}
\hline
Backbone                         & Method & HCE  & AP   & AP$^{50}$ & AP$^{75}$ & AP$^{S}$ & AP$^{M}$ & AP$^{L}$ \\
\hline
\multirow{6}{*}{ResNet-50-FPN}   & \multirow{2}{*}{FPN} & & 36.3 & 58.3 & 39.1 & 21.6 & 40.2 & 46.9 \\
                                 & & \checkmark & \textbf{38.4} & \textbf{61.0} & \textbf{41.8} & \textbf{22.9} & \textbf{42.5} & \textbf{49.1} \\ \cline{2-9}
                                 & \multirow{2}{*}{Mask R-CNN}  &             & 37.3 & 59.1 & 40.3 & 22.2 & 41.1 & 48.3 \\
                                 &  & \checkmark  & \textbf{38.8} & \textbf{61.3} & \textbf{42.1} & \textbf{23.2} & \textbf{42.8} & \textbf{49.7} \\ \cline{2-9}
                                 & \multirow{2}{*}{Cascade R-CNN} &         & 40.5 & 58.7 & 44.1 & 22.3 & 43.6 & 53.8 \\
                                 & & \checkmark & \textbf{41.7} & \textbf{60.5} & \textbf{45.0} & \textbf{23.4} & \textbf{44.9} & \textbf{55.2} \\
\hline
\multirow{6}{*}{ResNet-101-FPN}  & \multirow{2}{*}{FPN}   &      & 38.3 & 60.1 & 41.7 & 22.8 & 42.8 & 49.8 \\
                                 &  & \checkmark & \textbf{40.0} & \textbf{62.3} & \textbf{43.4} & \textbf{24.0} & \textbf{44.1} & \textbf{51.9} \\ \cline{2-9}
                                 & \multirow{2}{*}{Mask R-CNN} &           & 39.4 & 60.9 & 43.0 & 23.3 & 43.7 & 51.5 \\
                                 &  & \checkmark & \textbf{40.5} & \textbf{62.6} & \textbf{44.0} & \textbf{24.4} & \textbf{44.5} & \textbf{53.4} \\ \cline{2-9}
                                 & \multirow{2}{*}{Cascade R-CNN} & & 41.9  & 60.1 & 45.7 & 23.2 & 45.9 & 56.2 \\
                                 &   & \checkmark & \textbf{43.0} & \textbf{61.6} & \textbf{46.9} & \textbf{24.6} & \textbf{46.6} & \textbf{57.4} \\
\hline
\end{tabular}
\label{table:baseline}
\end{table}

To demonstrate the generality of our HCE framework, we consider three well-known region-based object detectors as our baseline systems, including Feature Pyramid Network (FPN)~\cite{fpn}, Mask R-CNN~\cite{maskrcnn} and Cascade R-CNN~\cite{cascade}. All detectors are instantiated with two different backbones, \ie, ResNet-50 and ResNet-101 with FPN. Integrating our framework with Mask R-CNN and Cascade R-CNN is as straightforward as with FPN. For example, we apply our framework within each training stage of Cascade R-CNN.

Comparison results on MS-COCO 2017 \texttt{val} are shown in Table~\ref{table:baseline}. Our HCE framework achieves consistent accuracy gains overall all baseline detectors on all evaluation metrics. Specifically, without the bells and whistles, our method improves 2.1\% and 1.7\% AP for FPN with ResNet-50 and ResNet-101 backbones, respectively. While for more advanced Mask R-CNN and Cascade R-CNN, our method also brings more than 1\% AP improvement on both ResNet-50 and ResNet-101 backbones, \textit{e.g.,} improving the AP for Mask R-CNN with ResNet-50-FPN from 37.3\% to 38.8\%.

Additionally, it can be observed that our improvements for Mask R-CNN and Cascade R-CNN baselines are not as significant as FPN. We conjecture that this is because Mask R-CNN and Cascade R-CNN themselves integrate mechanisms for better feature learning, which might overlap with the performance gains with our method. Specifically, Mask R-CNN benefits from extra accurate instance-level mask supervisions, while Cascade R-CNN enjoys IoU-specific multi-stage training to progressively refine object proposals and learn discriminative features for IoU-specific proposals. However, even in these cases, our method can also obtain $+1\%$ AP improvement over these competing baseline methods.

\subsection{Error Analyses}

\begin{figure}[t!]
	\centering
	\includegraphics[width=1.0\textwidth]{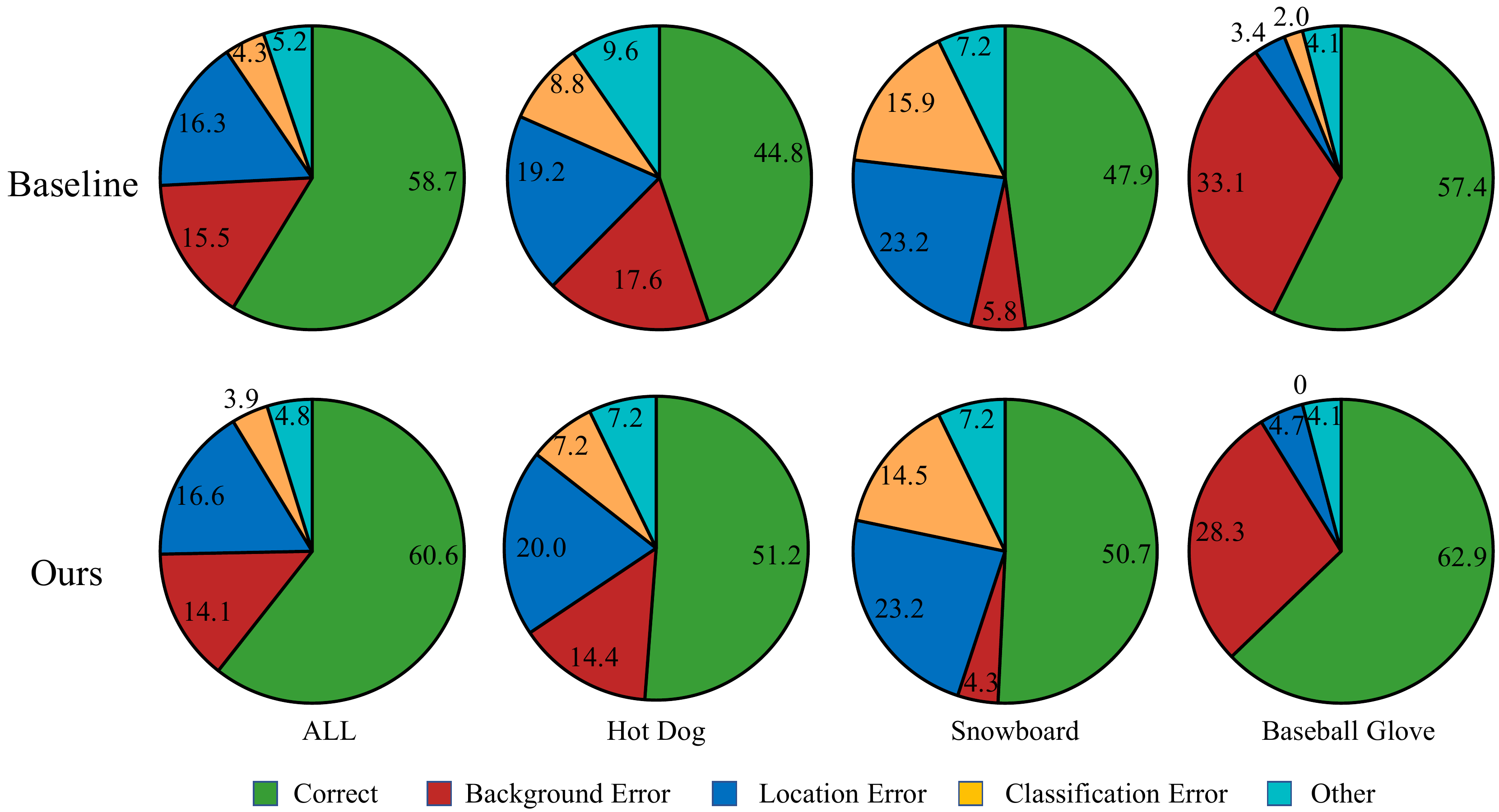}
	\caption{Error Analyses: These illustrations show the percentage of different error types in the top N detections (N = \# objects in that category).}
	\label{fig:error}
\end{figure}

In the following, we perform error analyses to further understand in what aspects our HCE framework improves the region-based object detectors.
Following the settings of~\cite{yolo}, we choose the top N predictions for each category during inference time. Each prediction is classified based on the type of error:
\begin{itemize}
	\item Correct: correct class and IOU $>$ 0.5
	\item Location Error: correct class and 0.1 $<$ IOU $<$ 0.5
	\item Background Error: IOU $<$ 0.1 for any object
	\item Classification Error: class is wrong and IOU $>$ 0.5
	\item Other: class is wrong and 0.1 $<$ IOU $<$ 0.5
\end{itemize}

We compare different error types between the FPN baseline and our method with ResNet-50 as backbone on MS-COCO 2017 \texttt{val}. Fig.~\ref{fig:error} shows the results of each error type averaged across all 80 categories, and each error type for ``\texttt{hot dog}'', ``\texttt{snowboard}'' and ``\texttt{baseball glove}'' which are highly dependent on context information. Obviously, our method can effectively improve the classification ability of region-based detector and reduce the background errors to a large extent, without compromising the localization performance or increasing other type of errors. Our improvements are particularly noticeable for context-dependent object categories. For example, the (normalized) correctly recognized instances of ``\texttt{hot dog}'' increase from 44.8\% to 51.2\%, while the background false positive detections reduce from 17.6\% to 14.4\%. These observations validate that our HCE framework can indeed improve the classification ability.

\subsection{Ablation Studies}
In this section, we conduct three series of ablation experiments to analyze the proposed method, using ResNet-50 as backbone on MS-COCO 2017 \texttt{val}.

\subsubsection{Context Embedding Operations}
We first investigate the impacts of different context embedding operations in our HCE framework. Specifically, there are three context embedding operations involved in our framework. Firstly, the image-level categorical embedding module employs multi-label learning (denoted as ``MLL'') to embed global-level context to advance the learning of context-dependent categories. Then, for further improving region-based classification, both the context-embedded instance-level feature (denoted as ``Instance'') and the context-aggregated global-level feature (denoted as ``Global'') are combined to generate hierarchical contextual RoI feature.

Table~\ref{table:context} shows the performance improvements by progressively integrating more context embedding operations. Solely applying MLL on the detection backbone gives $0.5\%$ AP improvement. This verifies that image-level categorical embedding advances the feature learning for context-dependent object categories. 
Then, the context-embedded instance-level feature which can be directly utilized by the detection head brings another $1.0\%$ AP improvement. Finally, global-level context embedding for contextual RoI feature improves $0.6\%$ AP. These results suggest that the context embedding operations in our framework are complementary with each other.

\begin{table}[t]
\centering
\setlength{\tabcolsep}{2.8pt}
\caption{Impacts of different context embedding operations on MS-COCO 2017 \texttt{val}. ``MLL'' means we leverage the image-level categorical embedding module to advance the learning of context-dependent categories. ``Instance'' and ``Global'' denotes that we utilize instance-level (cf. Eq~(\ref{eq:instance})) or global-level (cf. Eq~(\ref{eq:global})) contextual features to further improve the region-wise detection head.}
\begin{tabular}{c|ccc|ccc|ccc}
\hline
Method                        & MLL & Instance    & Global    & AP   & AP$^{50}$ & AP$^{75}$ & AP$^{S}$  & AP$^{M}$ & AP$^{L}$ \\
\hline
\multirow{4}{*}{FPN} &             &            &            & 36.3 & 58.3 & 39.1 & 21.6 & 40.2 & 46.9 \\
                                  & \checkmark  &            &            & 36.8 & 58.9 & 39.7 & 21.9 & 40.5 & 47.2 \\
                                  & \checkmark  & \checkmark &            & 37.8 & 59.9 & 40.9 & 22.2 & 41.4 & 48.9 \\
                                  & \checkmark  & \checkmark & \checkmark & \textbf{38.4} & \textbf{61.0} & \textbf{41.8} & \textbf{22.9} & \textbf{42.5} & \textbf{49.1} \\
\hline
\end{tabular}
\label{table:context}
\end{table}

\begin{table}[t]
\centering
\setlength{\tabcolsep}{5pt}
\caption{Effects of different fusion strategies during \textit{training}, evaluated by detection performance on MS-COCO 2017 \texttt{val}. The models share the same backbone network ResNet50-FPN. ``FF Train'' means that we apply feature fusion (FF) for training, while ``CF Train'' means confidence fusion (CF) are applied for training.}
\begin{tabular}{c|cc|ccc|ccc}
\hline
Method                            & FF Train   & CF Train & AP   & AP$^{50}$ & AP$^{75}$ & AP$^{S}$  & AP$^{M}$ & AP$^{L}$ \\
\hline
\multirow{4}{*}{FPN} &            &             & 36.8 & 58.9 & 39.7 & 21.9 & 40.5 & 47.2 \\
                                  & \checkmark &             &  37.6 & 60.3 & 40.7 & 22.5 & 41.4 & 48.2\\
                                  &            & \checkmark  &  
                                  37.4 & 60.2 & 40.1 & 23.0 & 41.1 & 47.6\\
                                  & \checkmark & \checkmark  & \textbf{38.4} & \textbf{61.0} & \textbf{41.8} & \textbf{22.9} & \textbf{42.5} & \textbf{49.1} \\
\hline
\end{tabular}
\label{table:head}
\end{table}

\begin{table}[t]
\centering
\setlength{\tabcolsep}{4pt}
\caption{Effects of different fusion strategies in testing, which are evaluated by the inference time and detection performance on MS-COCO 2017 \texttt{val}. Note that all models are trained with both fusion strategies. ``FF Test'' denotes that we evaluate the feature fusion (FF) strategy during inference, while ``CF Test'' means the results are evaluated by confidence fusion (CF) strategy. Inference speed is evaluated on a single 1080ti GPU.}
\begin{tabular}{c|cc|c|ccc|ccc}
\hline
Method                            & FF Test & CF Test & Speed & AP   & AP$^{50}$ & AP$^{75}$ & AP$^{S}$ & AP$^{M}$ & AP$^{L}$ \\
\hline
\multirow{4}{*}{FPN}  &                  &                & 0.087s & 36.3 & 58.3 & 39.1 & 21.6 & 40.2 & 46.9 \\
                                   & \checkmark       &                & 0.090s & 38.2 & 60.8 & 41.5 & 22.6 & 42.2 & 49.0 \\
                                   &                  & \checkmark     & 0.094s & 38.3 & 60.8 & 41.6 & 22.8 & 42.3 & 49.0 \\
                                   & \checkmark       & \checkmark     & 0.100s & \textbf{38.4} & \textbf{61.0} & \textbf{41.8} & \textbf{22.9} & \textbf{42.5} & \textbf{49.1} \\ 
\hline
\end{tabular}
\label{table:test}
\end{table}

\subsubsection{Fusion Strategies in Training}
We consider the proposed two fusion strategies, feature fusion and confidence fusion, are complementary to each other. To verify this, we evaluate the performance by training the model with feature fusion and confidence fusion individually, as well as both of them. Table~\ref{table:head} shows the results of different fusion strategies. Specifically, ``FF Train" means that we apply feature fusion (FF) for training, while ``CF Train" means confidence fusion (CF) are applied for training. Utilizing feature fusion and confidence fusion individually for training can outperform the baseline (FPN with MLL) by $0.8\%$ and $0.6\%$ AP, respectively. Training with both fusion strategies achieves the best result, and is clearly better than using each individual fusion strategy separately.

\subsubsection{Fusion Strategies in Testing}
We also evaluate each fusion strategy independently during inference, with all HCE models trained with both fusion strategies. Table~\ref{table:test} shows the results of each fusion strategy and the combined fusion strategies. ``FF Test'' denotes that we evaluate the feature fusion (FF) strategy during inference, while ``CF Test'' means that the results are evaluated by confidence fusion (CF) strategy. We can see that once the model is trained with both fusion strategies, using only one fusion branch for inference will not cause obvious accuracy drop, but brings computational economy. For example, using the feature fusion branch for inference adds very minimal time cost (0.003s) to the baseline, but increases the AP from 36.3\% to 38.2\%. These results also prove the complementarity of the proposed two fusion strategies.

\begin{table}[t]
\centering
\setlength{\tabcolsep}{1pt}
\caption{Comparisons with the state-of-the-art single-model detectors on MSCOCO 2017 \texttt{test-dev}. ``*'' denotes using tricks (with bells and whistles) during inference.}
\begin{tabular}{c|c|cccccc}
\hline
Method                                           & Backbone                 & AP   & AP$^{50}$ & AP$^{75}$ & AP$^{S}$  & AP$^{M}$ & AP$^{L}$ \\
\hline
YOLOv3~\cite{yolov3}                             & Darknet-53               & 33.0 & 57.9 & 34.4 & 18.3 & 35.4 & 41.9 \\
SSD513~\cite{ssd}                                & Res101                   & 31.2 & 50.4 & 33.3 & 10.2 & 34.5 & 49.8 \\
RetinaNet~\cite{focalloss}                       & Res101-FPN               & 39.1 & 59.1 & 42.3 & 21.8 & 42.7 & 50.2 \\
FCOS~\cite{fcos}                                 & Res101-FPN               & 41.5 & 60.7 & 45.0 & 24.4 & 44.8 & 51.6 \\
\hline
FPN~\cite{fpn}                                   & Res101-FPN               & 36.2 & 59.1 & 39.0 & 18.2 & 39.0 & 48.2 \\
Mask R-CNN~\cite{maskrcnn}                       & Res101-FPN               & 38.2 & 60.3 & 41.7 & 20.1 & 41.1 & 50.2 \\
Cascade R-CNN~\cite{cascade}                     & Res101-FPN               & 42.8 & 62.1 & 46.3 & 23.7 & 45.5 & 55.2 \\
Deformable R-FCN*~\cite{deformable}              & Aligned-Inception-ResNet & 37.5 & 58.0 & 40.8 & 19.4 & 40.1 & 52.5 \\
DCNv2*~\cite{dv2}                                & Res101-DeformableV2      & 46.0 & \textbf{67.9} & \textbf{50.8} & \textbf{27.8} & 49.1 & \textbf{59.5} \\
IoU-Net~\cite{iounet}                            & Res101-FPN               & 40.6 & 59.0 & --   & --   & --   & --   \\
TridentNet~\cite{trident}                        & Res101                   & 42.7 & 63.6 & 46.5 & 23.9 & 46.6 & 56.6 \\
Cascade +Rank-NMS~\cite{ranking}                 & Res101-FPN               & 43.2 & 61.8 & 47.0 & 24.6 & 46.2 & 55.4 \\
\hline
HCE FPN                                      & Res101-FPN               & 41.0 & 63.5 & 44.7 & 23.4 & 44.2 & 52.2 \\
HCE Mask R-CNN                               & Res101-FPN               & 41.6 & 63.9 & 45.4 & 23.7 & 44.7 & 53.1 \\
HCE Cascade R-CNN                            & Res101-FPN               & 44.1 & 63.2 & 47.9 & 25.2 & 46.9 & 57.0 \\
HCE Cascade R-CNN*                           & Res101-FPN               & \textbf{46.5} & 65.6 & 50.6 & 27.4 & \textbf{49.9} & 59.4 \\
\hline
\end{tabular}
\label{table:soa}
\end{table}

\subsection{Comparisons with State-of-the-art}
We compare our proposed method with state-of-the-art on MS-COCO 2017 \texttt{test-dev}. For fair comparisons, we report the performance of all methods with single-model inference. Specifically, we apply our method on FPN, Mask R-CNN and Cascade R-CNN in 2$\times$ training scheme without bells and whistles. Table~\ref{table:soa} shows all comparison results. 

Our hierarchical context embedding framework, when integrated with FPN, Mask R-CNN and Cascade R-CNN object detectors, consistently outperforms state-of-the-art object detectors using the same backbone network. For fairly comparisons with Deformable R-FCN* and DCNv2* which adopt multi-scale 3x training scheme and multi-scale testing, we follow the same experimental setting to train our HCE Cascade R-CNN*. It gives an AP of 46.5\%, which surpasses R-FCN* and DCNv2*. These results demonstrate the superior performance of the proposed context embedding framework.

\section{Conclusions}

In this paper, we investigated the limitation of context information on conventional region-based detectors, and proposed a novel and effective Hierarchical Context Embedding (HCE) framework to facilitate the classification ability of current region-based detectors. Comprehensive experiments demonstrated the consistent outperforming accuracy on almost all existing mainstream region-based detectors, include FPN, Mask R-CNN and Cascade R-CNN. In the future, we will concentrate in extending the usage scope of our HCE framework and adapting it to one-stage detection paradigm.\\

\noindent
\textbf{Acknowledgements}: 
Z.-M. Chen's contribution was made when he was an intern in Megvii Research Nanjing. This research was supported by the National Key Research and Development Program of China under Grant 2017YFA0700800, the National Natural Science Foundation of China under Grants 61772257 and  the Fundamental Research Funds for the Central Universities 020914380080.

\clearpage
%
%
\bibliographystyle{splncs04}
\bibliography{egbib}
\end{document}